\documentclass[11pt]{article}

\usepackage[final]{acl}

\usepackage{times}
\usepackage{latexsym}

\usepackage[T1]{fontenc}

\usepackage[utf8]{inputenc}

\usepackage{microtype}
\usepackage{lipsum}
\usepackage{hyperref}

\usepackage{inconsolata}
\usepackage{amsmath}
\usepackage{amsfonts}
\usepackage{stmaryrd}
\usepackage{float}
\usepackage{booktabs} 
\usepackage{multirow}

\usepackage{footnote}
\usepackage{caption}
\usepackage{subcaption} 
\usepackage{graphicx}
\usepackage{tikz}
\usetikzlibrary{arrows.meta, patterns, shapes, positioning, calc}
\usepackage{pgfplots}
\pgfplotsset{compat=1.18}

\newcommand{\bucket}[2]{\([#1,#2)\)}

 \title{Progress Ratio Embeddings: An Impatience Signal for Robust Length Control in Neural Text Generation}
\author{
  Ivanhoé Botcazou \qquad
  Tassadit Amghar \qquad
  Sylvain Lamprier \qquad
  Frédéric Saubion  \\
  LERIA, University of Angers \\
  \texttt{\{ivanhoe.botcazou, tassadit.amghar, sylvain.lamprier,  frederic.saubion \}@univ-angers.fr}
}

\begin{document}
\maketitle
\begin{abstract}

Modern neural language models achieve high accuracy in text generation, yet precise control over generation length remains underdeveloped. In this paper, we first investigate a recent length control method based on \emph{Reverse Positional Embeddings (RPE)} and show its limits when control is requested beyond the training distribution. In particular, using a discrete countdown signal tied to the absolute remaining token count leads to instability. To provide robust length control, we introduce \emph{Progress Ratio Embeddings (PRE)}, as continuous embeddings tied to a trigonometric {\em impatience} signal. PRE integrates seamlessly into standard Transformer architectures, providing stable length fidelity without degrading text accuracy under standard evaluation metrics. We further show that PRE generalizes well to unseen target lengths. Experiments on two widely used news-summarization benchmarks and a popular question generation dataset validate these findings.\footnote{Code available at: \href{https://github.com/Ivanbtz9/ProgressRatioEmbeddings}{Ivanbtz9/ProgressRatioEmbeddings}}

\end{abstract}

\section{Introduction}

Over the past few years,  Language Models (LMs) have achieved remarkable progress in natural language processing, reaching state-of-the-art performance across a wide range of text generation tasks. Despite these advances, generating outputs that meet specific structural or stylistic constraints remains challenging. Among these constraints, length control is particularly fundamental, as the appropriate degree of compression or elaboration often depends on the task and the desired granularity of the information. In tasks such as summarization, for instance, the compression ratio between the source and the output naturally correlates with the number of words or tokens produced. Yet, controlling this aspect precisely remains both a technical and a cognitive challenge for neural language models \citep{kikuchiControllingOutputLength2016a}.

During inference, major LMs generate tokens sequentially until the process terminates by the sampling of the \texttt{<EOS>} token. However, this stopping criterion is inherently stochastic, and the decision mechanism guiding token selection at each step depends primarily on the previously generated context. As a result, the overall planning of the generation process lacks transparency, offering only limited opportunities for explicit control. In practice, users can request outputs of varying lengths, ranging from concise phrases to long, elaborated texts. For example, in abstractive summarization, onecan look for either very short TL;DR-style summaries or slightly compressed paragraphs that preserve much of the original content. While instruction-following Large Language Models (LLMs) can express length preferences through prompting \citep{Lou2024,dong-etal-2024-survey}, this form of in-context control remains restricted. It is effective only for very large models trained on extensive amounts of general-purpose data. In this paper, we focus on the setting of Small Language Models (SMLs) trained on restricted, domain-specific datasets, where length control must instead be achieved through architectural design.

Recent approaches have addressed the problem of target length control in text generation, through the use of length-informed positional embeddings integrated into the inputs fed to the decoder of the LM. For example, \citep{miculicichREPILOTSummarizationPrecise2023, Butcher_2025}  propose to feed the Transformer decoder with reverse positional encodings that depend on the remaining number of tokens to be decoded given a desired target length. Although this enables alright length control for summaries whose lengths fall within a restricted window centered on the mode of the training distribution, in this paper, we show that it performs poorly for out-of-distribution target lengths.  

To avoid this limitation, we introduce continuous \emph{Progress Ratio Embeddings} (PRE), which reduce the dependence on the discrete target length provided to the model. In contrast to discrete Reverse Positional Embeddings (RPE), PRE offers a smooth and continuous representation of progress that generalizes beyond the training distribution of reference lengths. PRE embeddings are introduced through trigonometric impatience signals whose frequency gradually increases until the end of generation. This design aligns the decoding process more closely with the user-specified target length and, by construction, makes it less dependent on the distribution of training reference lengths. We further integrate PRE into two notable pre-trained Encoder-Decoder architectures \citep{NIPS2017_3f5ee243}, achieving robust length invariance without compromising relevance or informativeness in neural text generation tasks.  

Our main contributions are threefold:

\begin{enumerate}
    \item[\,(1)\,] We propose a novel length control strategy, called Progress Ratio Embeddings (PRE), based on continuous “impatience” signals that modulate the decoding process according to the completion progress;
    
    \item[\,(2)\,] We reproduce and adapt the Reverse Positional Embeddings (RPE) method on a classical encoder--decoder architecture, and provide an analysis of its strengths and weaknesses;   
    
    \item[\,(3)\,] 
    We demonstrate, on standard text summarization, that our PRE model can generalize length control beyond the training distribution, producing outputs of previously unseen lengths. 
    
\end{enumerate}

The paper is organized as follows. Section \ref{sec:related_work} presents related works. In Section \ref{sec:cont_len_cont} we introduce our method, called PRE. Section \ref{sec:experimental_setup} describes the setup that is used for the experiments presented in Section \ref{sec:rpe_results_analysis} and Section \ref{sec:pre_results_analysis}.

\section{Related Work}
\label{sec:related_work}

The large number of parameters and the probabilistic nature of LMs associate them with Black Box algorithms, which suffer from well-known weaknesses such as a lack of explicability and limited controllability in planning \citep{zhao2023explainabilitylargelanguagemodels}. 
Given an input context, the vast space of possible continuations forces an autoregressive model to select a trajectory via sampling to maximize a context-conditioned probability.  

To improve controllability, recent research has shown that LM outputs can be empirically enhanced through \textit{chain-of-thought prompting} (CTP). 
Within this perspective, existing study \citep{jiePromptBasedLengthControlled2024} propose prompt-based methods combined with reinforcement learning to train prompt extractors that enhance both performance and controllability. 
Moreover, by integrating prompting with iterative trajectory sampling under a Metropolis--Hastings framework, \citet{gu2024lengthcontrolledgenerationblackbox} demonstrate how LLMs can be regulated to enforce constraints such as output length, without requiring parameter modifications. In this paper, we instead focus on smaller language 
models trained on domain-specific or resource-constrained datasets, where reference contents are distributed within a narrow length range. In such constrained settings, where contextual information about length is limited and prompting is not feasible, explicit architectural mechanisms become necessary to guarantee effective and reliable length control.  

An alternative line of research focuses on modifying the decoding process itself. 
Decoding-based approaches, such as \textit{LenAtten} \citep{yuLenAttenEffectiveLength2021}, extend the autoregressive generation mechanism by conditioning the prediction head not only on the decoder’s hidden state but also on an additional length-context vector, thereby enabling finer-grained control over the generated output length. Beyond prompting and decoding, other studies introduce structural modifications to the model architecture. For example, the Length-Aware Attention Mechanism (LAAM), \citep{liuLAAMLengthControl2022} operates at the attention level by gradually adjusting the cross-attention weights in a length-dependent manner, particularly between the \texttt{<EOS>} token in the input text and each decoder token (Appendix~\ref{sec:appendixE}). 

In parallel, the remaining length is encoded via specialized positional representations, following the control mechanism introduced by \citet{takasePositionalEncodingControl2019} (Appendix~\ref{sec:appendixG}). Based on this concept, \citet{miculicichREPILOTSummarizationPrecise2023} and \citet{Butcher_2025} fine-tune pre-trained Transformers to incorporate RPE embeddings. 
At each decoding step, a discrete countdown index representing the remaining length is converted into a vector of the same dimension as the token and positional embeddings, and injected into the decoder (Appendix~\ref{sec:appendixB}). By training through next-token prediction, this additional signal enables the model to predict more accurately the expected stopping point.  

However, the RPE method is explicitly dependent on the target length observed during training. 
As shown in Section~\ref{sec:rpe_results_analysis}, this dependency constitutes a critical limitation, RPE achieves correct control only within the training length distribution, but globally fails to generalize for out-of-distribution lengths. This weakness highlights the need for novel architectural mechanisms that enable deterministic and more invariant length control in text generation. To address this challenge, we introduce the PRE method.

\section{Continuous Length Control}
\label{sec:cont_len_cont}

In this section, we provide a brief overview of the role of positional embeddings in Transformer models.
We then introduce our PRE method, which injects a lightweight {\em impatience signal} into the decoder inputs of a vanilla Transformer. We justify the theoretical foundation from the digital signal processing and expose the global fine-tuning aspects. 

\subsection{Transformers needs Positional Information}

Transformer-based LMs process sequences of ordered tokens, where each token $t$ is mapped to a token embedding\footnote{An embedding is a  dense vector of dimension $d_{\text{model}}$} $E_t \in \mathbb{R}^{d_{\text{model}}}$. Because of the self-attention setup by itself, the input tokens are permutation invariant, the model must be given positional information. Each token index position can be mapped to a positional embedding 
$P_t \in \mathbb{R}^{d_{\text{model}}}$, which is added to the token 
embedding $E_t$ before the first Transformer layer :
\begin{equation}
\label{eq:pos_emb}
X_t = E_t + P_t.
\end{equation}

This addition places content and position information in the same vector $X_t$ and allows attention to condition on where a token occurs, not only on what it is. Depending on the model configuration, positional information can be encoded with fixed sinusoidal embeddings as initially introduced in \citep{NIPS2017_3f5ee243}. Alternatively, postional embeddings can be learned during the pre-training as presented in \citep{gehring2017convolutionalsequencesequencelearning, radfordImprovingLanguageUnderstanding}. Some models incorporate relative positional embeddings that capture relative distance between sequence elements \citep{shaw-etal-2018-self}. Modern LLMs typically use Rotary Positional Embeddings (RoPE), which are relative positional encodings applied inside each attention block \citep{su2023roformerenhancedtransformerrotary}. In all cases, without positional information, any permutation of the input tokens yields the same multiset of attention scores. 

\begin{figure}[H] 
    \centering
    \includegraphics[width=0.4\textwidth]{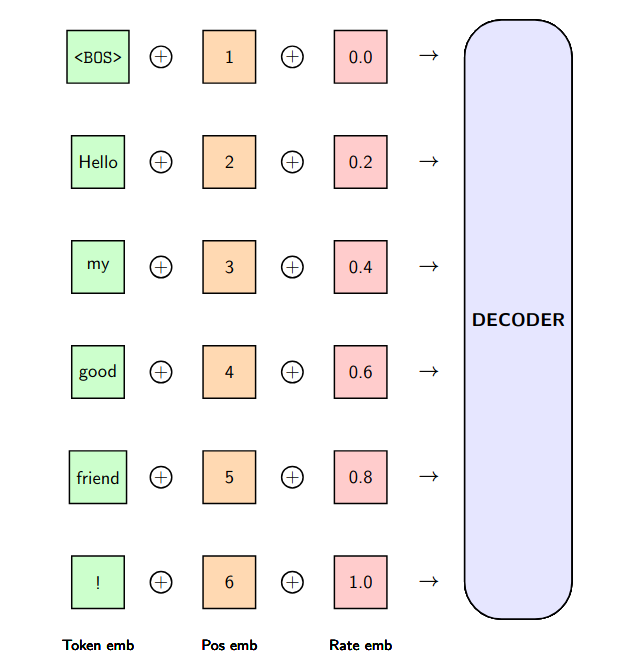} 
    \caption{Illustration of Progress Ratio Embeddings added to token and positional embeddings to form the input of the decoder. } 

    \label{fig:PRE_schema}
\end{figure}

 At the decoding step, the RPE methods additionally provide new position embeddings in reverse order to indicate how many tokens remain to be generated, rather than only giving the absolute position from the start of generation. When the model is trained with such reversed embeddings counting down from the initial target length to zero over the training reference the decoder block can modulate generation length at inference time according to a user-specified target length. However, we argue that this approach, while effective for lengths seen during training, does not generalize well to unseen target lengths, which are often required in practical applications. In the following, we propose a new method to overcome this limitation.

\subsection{Progress Ratio Embeddings as an Impatience Signal}
\label{subsec:PRE_methode}

We introduce the PRE method based on progress ratios \( r \in [0,1] \) that quantify the advancement of the decoding process. At decoding step \(t\) for a requested target length \(l\), we can define  a progress ratio: \( r_{t} = \frac{t}{l} \). As illustrated in Figure~\ref{fig:PRE_schema}, a ratio-dependent embedding is added  to the inital decoder input embeddings :
\begin{equation}
\label{eq:pos_rpe_emb}
X_{t} \;=\; E_t \;+\; P_t \;+\; \xi\!\left(r_{t}\right)
\end{equation}
where \( \xi : [0,1] \to \mathbb{R}^{d_{\text{model}}} \) denotes what we call the \emph{Progress Ratio Embdedding function}. 
A crucial advantage over the finite set of reversed positional embeddings used in RPE process is that PRE defines embeddings continuously for any ratio within the interval \([0,1]\). 
This allows the model to compute length-control information on the fly, covering the entire progress range rather than being restricted to discrete positions seen during training.

Our goal is to construct an embedding $\xi(r_t)$ that is rich enough to avoid overfitting to the specific ratios seen during training,  that may differ at inference time when new target lengths are requested. We therefore seek a smooth representation that forms a meaningful metric space over the range $[0;1]$. Specifically, we assume that for any possible ratio \( r \in [0,1] \), there exists a regular periodic function that represents the \emph{impatience signal} of the process. By analogy, this signal can be seen as the impatience of a listener awaiting the end of a message. Building on signal sampling theory, we design a stress signal with pulsation $\omega_r$ (i.e., corresponding to a frequency $F=\omega_r/(2\pi)$) that increases with the progress ratio $r$, which enables the decoder to receive embeddings that encode the generation progress across $[0,1]$. We consider in the following $\omega_r=r \times M$, with $M$ an hyper-parameter that scales the frequency increase of the signal  (Appendix~\ref{sec:appendixA}).  

The Nyquist–Shannon theorem \citep{1697831} guarantees that any periodic, band-limited function can be perfectly reconstructed from samples taken at a rate $F_s$ exceeding twice its highest frequency component $F_{max}$.  Furthermore, sinusoids provide a natural basis for such signals, the use of sine and cosine components ensure continuous phase encoding. Following this principle, we design our embedding so that each \emph{pair} of dimensions encodes a single frequency component via cosine and sine in alternating dimensions. Formally, for \(1 \leq j \leq d_{\text{model}}\),
\begin{equation}
\label{eq:pre_def}
\xi(r)_j =
\begin{cases}
\sin\!\left( \omega_r \cdot x_j  \right), & \text{if $j$ is even}, \\[2pt]
\cos\!\left( \omega_r \cdot x_j \right), & \text{if $j$ is odd},
\end{cases}
\end{equation}
where \(\omega_r\) controls the frequency of oscillations given by the progress ratio $r$. This formulation uses  $x_j=\frac{2 \,  \lfloor j/2 \rfloor }{d_{\text{model}}}$ as the independent variable for each dimension $j$, which  corresponds to a sampling rate of \(F_s = d_{\text{model}}/2\) given points regularly spaced over \([0,1]\). Taking $M=d_{model} \cdot \frac{\pi}{2}$ allows to benefit from the full capacity of the considered representation, while ensuring that the Nyquist–Shannon condition  is satisfied,  since we have in that case: 
$F = \frac{\omega_r}{2\pi} \leq 1\cdot  d_{model} /4 = \frac{d_{\text{model}}}{4}$,  and thus: $F_s = \frac{d_{\text{model}}}{2} =   2F_{max}$.  In practice, we use \( M = \frac{d_{\text{model}}}{2} \), leaving a small safety margin below the Nyquist bound to accommodate rounding and numerical precision limits. 

\subsection{Finetuning with PRE}

Let $A := (A_0, \ldots, A_n)$ denote the sequence of tokens corresponding to the context of the task (e.g., the source content), and $S := (S_0, \ldots, S_l)$ the sequence of tokens in a \emph{"gold"}  reference (e.g., a target text), where $l$ is the expected number of tokens in the produced text. Following Section~\ref{subsec:PRE_methode}, we consider an associated family of Progress Ratio Embeddings $\Xi = (\xi(r_0),\ldots,\xi(r_l)) $ related to the expected target length $l$. Given a pretrained  model\footnote{In our experiments, we focus on encoder-decoder models, although our technique can be straightforwardly applied to decoder-only architectures by initializing the progress ratios at the beginning of the expected sequence S.}, our objective is to adapt its parameters so that it predicts the conditional probability $P(S \mid A , l)$.

Following the maximum likelihood estimation (MLE) principle, we optimize the model parameters $\theta$ to maximize the product of conditional probabilities. This product can be factorized using the chain rule ($\star$) and the causal property ($\star\star$) of the decoder:
\begin{align*}
 P_\theta(S \mid A,\Xi) &\overset{\star}{=} \prod_{t=0}^{l} P_\theta\!\left(S_{t}\,\middle|\,S_{<t}, A, \Xi\right)\\
 &\overset{\star\star}{=}  \prod_{t=0}^{l} P_\theta\!\left(S_{t}\,\middle|\,S_{<t}, A, \Xi_{< t}\right).
\end{align*}

The training objective is defined by the standard cross-entropy loss, applied over mini-batches of size $m$\footnote{Computational resources allowed us to use $m=5$}. We minimize the empirical loss over a batch 
$\mathcal{B} = \{(A^i, S^i, \Xi^i)\}_{1\leq i \leq m}$, 
defined as the average negative log-likelihood across all tokens. 

\begin{align*}
\mathcal{L}_{\mathcal{B}}(\theta) 
&= - \frac{1}{m} \sum_{i=1}^{m} \sum_{t=0}^{l_i} 
\log P_\theta\!\left(S^i_t \,\middle|\, S^i_{<t}, A^i, \Xi^i_{<t}\right)
\end{align*}

Minimizing the sequence loss while conditioning the decoder on a normalized progress ratio provides a consistent source of temporal side information. To enhance generalization, Gaussian noise is injected into each ratio (Eq. \ref{eq:gaussian_noise}), exposing the model to a continuous spectrum of values over [0,1] and promoting smooth interpolation rather than dependence on a few discrete targets. 

\begin{equation}
    \label{eq:gaussian_noise}
    \begin{aligned}
    &r = Clip\left(r + \dfrac{2\delta}{d_{\text{model}}} ~; ~0 ~;~ 1\right) \quad\\
    &\text{where} \quad \delta \sim \mathcal{N}(0,1)
\end{aligned}
\end{equation}

Once the generated sequence reaches or exceeds the target length, the ratio is clipped at 1 causing the impatience signal to saturate and thereby discouraging further generation.

\section{Experimental Setup}
\label{sec:experimental_setup}

In this section, we present and justify our choices of model, datasets, hyperparameters, and evaluation metrics as integral components of the PRE method’s implementation and evaluation. 

\subsection{Model}

For our experimentations, we used the proven BART encoder-decoder model as backbone \citep{lewis2019bartdenoisingsequencetosequencepretraining}, it achieves strong performances on specific tasks and remains highly suitable for fine-tuning. As a sequence-to-sequence model pre-trained with a denoising objective inspired by \textsc{BERT} \citep{devlin2019bertpretrainingdeepbidirectional}, it has been shown to be effective in a range of NLP applications, including abstractive summarization and question generation task. Released by \citet{metaai} with about 400M parameters\footnote{We use the BART large version, where $d_{\text{model}} = 1,024$}, it can be trained on standard computing resources and supports inputs of up to 1,024 tokens.

\subsection{Datasets}
We conduct experiments on two widely used news-summarization benchmarks and on a question-answering dataset.

\noindent \textbf{CNN/DM} \citep{nallapati-etal-2016-abstractive} contains articles paired with summaries that are typically closely aligned with source sentences. 

\noindent \textbf{XSum} \citep{narayan-etal-2018-dont} provides highly abstractive one-sentence summaries that require substantial rephrasing beyond the source text.

\noindent \textbf{SQuAD} \citep{rajpurkar2016squad100000questionsmachine}
is the Stanford Question Answering Dataset, consisting of questions posed on Wikipedia articles.

\subsection{Metrics}
For each input article, there may exist multiple human-judged acceptable candidates that preserve the same level of information. Due to the existence of these valid alternatives, identifying metrics that accurately capture the relevance and informativeness of a generated summary remains highly challenging \citep{Koh_2022}. 

Given an input text and a gold-standard output summary, the most common metrics used to evaluate the relevance of abstractive candidates are ROUGE-1, ROUGE-2, and ROUGE-L \citep{linROUGEPackageAutomatic}, which provide an F1-score based on $n$-gram overlap between the generated output and the reference summary. However, these metrics are often questioned, as they do not correlate strongly with human judgments \citep{deutschReExaminingSystemLevelCorrelations2022a}. Despite this limitation, ROUGE remains the primary score reported in a large number of studies, including ours. Acknowledging the weaknesses of ROUGE, the BERTScore (B.S.) \citep{zhangBERTScoreEvaluatingText2020} has been proposed to avoid this lack of relevance. To further extend our evaluation, we employ a fine-grained LLM-as-a-judge framework specifically tailored for summarization. This approach is inspired by the FineSurE evaluation methodology \citep{song2024finesurefinegrainedsummarizationevaluation}, with comprehensive implementation details provided in (Appendix~\ref{sec:appendixH}). We evaluate length controllability using the mean absolute error (MAE) between the generated and target lengths. Unless otherwise stated, lengths are measured in tokens as defined by each model’s tokenizer.

\subsection{Hyperparameters and Devices}
All models were trained on the full training set using the AdamW optimizer \citep{loshchilov2019decoupledweightdecayregularization}, with hyperparameters $\beta_1 = 0.9$, $\beta_2 = 0.99$, and $\epsilon = 1\times 10^{-8}$. The learning rate was set to $1\times 10^{-5}$. All reported experiments and results are evaluated on the full test split of each dataset.

\section{RPE results and limitations of the summary task}
\label{sec:rpe_results_analysis}

We begin by reproducing the RPE method with BART-L and fine-tuning it on two famous summary benchmark datasets. This reproduction confirms the reported length-control behavior, as shown in Table~\ref{tab:length_control}, consistently regarding results from the original paper \citep{miculicichREPILOTSummarizationPrecise2023}. It also indicates this method maintains strong summarization quality relative to baselines without length control, which we reported in Table~\ref{tab:scores_results}. RPE-BART-L achieves accurate length control within each 10-token target-length bucket.

\begin{figure}[t!]
    \includegraphics[width=0.46\textwidth]{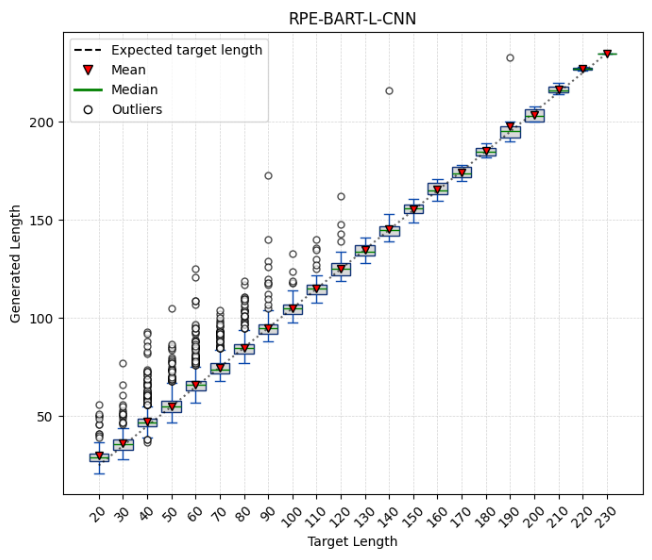}
    \caption{MAE by target-length bucket (10 tokens) for RPE-BART-L on CNN/DailyMail.}
    \label{fig:rpe_summaries}
\end{figure}

\begin{figure}[t!]
    \includegraphics[width=0.46\textwidth]{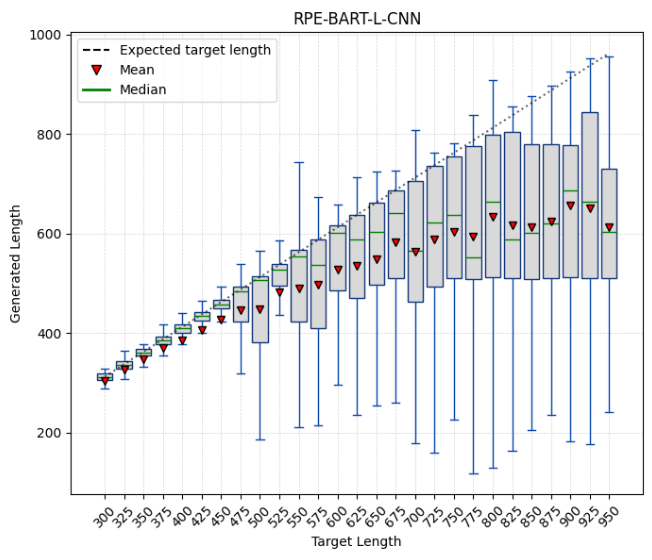}
    \caption{MAE by target-length bucket (25 tokens) for RPE-BART-L on CNN/DailyMail when target lengths above 300 tokens are requested.}
    \label{fig:rpe_long_summaries}
\end{figure}

However, in approximately $2\%$ of cases, the MAE between the generated and expected lengths exceeds $10$ tokens (see outliers in Figure~\ref{fig:rpe_summaries}). Moreover, as shown in Figure~\ref{fig:rpe_long_summaries}, when the requested target length exceeds $350$ tokens, the effectiveness of length control degrades markedly. \newpage This lack of generalization\footnote{Target lengths chosen according to the input article length and sampled at random from out-of-distribution lengths.} is precisely what motivated the introduction of our - length  invariant - PRE approach, that we experiment in the next section. 

\section{PRE results and analysis}
\label{sec:pre_results_analysis}

\subsection*{Summaries under length control}

\begin{figure}[t!]
    \includegraphics[width=0.45\textwidth]{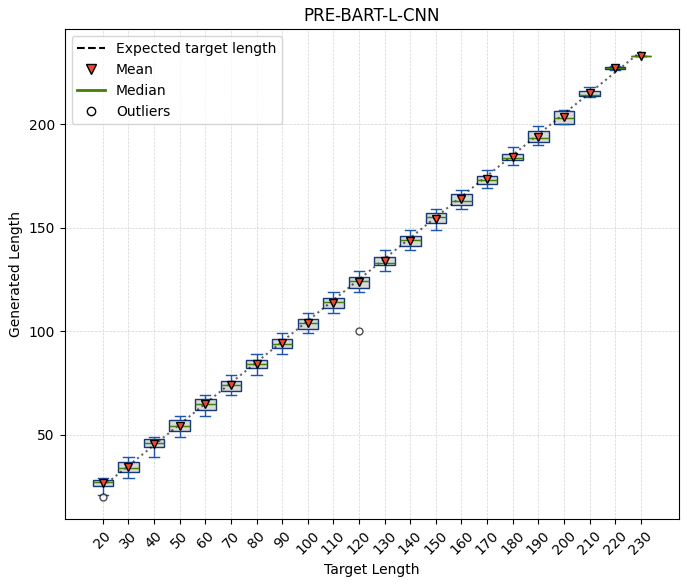}
    \caption{MAE by target-length bucket (10 tokens) for PRE-BART-L on CNN/DailyMail.}
    \label{fig:pre_summaries}
\end{figure}

Results reported in Table~\ref{tab:scores_results} demonstrate that our PRE method consistently yields high-quality summaries while maintaining precise length control. Specifically, the PRE approach sustains high ROUGE and BERTScore performance by achieving a low Mean Absolute Error (MAE) between target and generated lengths across the full spectrum of length constraints (see Figures~\ref{fig:pre_summaries} and \ref{fig:three-wide-cnn-length-distrib}). Table~\ref{tab:length_control} further contrasts PRE with both no-control baselines and our RPE reimplementation, confirming that PRE delivers superior controllability without sacrificing quality.

\begin{table}[h!]
    \centering
    \small
    \renewcommand{\arraystretch}{1.1}
    
    \begin{tabular}{@{}lccc@{}}
        \toprule
        \textbf{Dataset} & \textbf{Model} & \textbf{Type} & \textbf{MAE}$^{\scriptsize \downarrow}$ $\pm$ \textbf{SD} \\ 
        \midrule
        \multirow{4}{*}{\textbf{CNN/DM}} 
            & BART-L & N & 19.2 $\pm$ 17 \\
            & RPE-BART-L & G & 1.6 $\pm$ 3.6 \\
            & PRE-BART-L & G & 0.5 $\pm$ 0.3 \\ 
        \midrule
        \multirow{4}{*}{\textbf{XSum}} 
            & BART-L & N & 5.8 $\pm$ 5 \\
            & RPE-BART-L & G & 0.7 $\pm$ 1.1 \\
            & PRE-BART-L & G & 0.1 $\pm$ 0.2 \\ 
        \bottomrule
    \end{tabular}
    \caption{Length control evaluation on CNN/DailyMail and XSum using MAE and standard deviation (SD). \textbf{Legend:} G = reference length; N = no explicit length control. Lower values indicate better control.}
    \label{tab:length_control}
\end{table}

\begin{table}[H]
\centering
\small
\setlength{\tabcolsep}{1pt}
\renewcommand{\arraystretch}{1.1}

\begin{tabular}{lccccc}
\toprule
\multicolumn{6}{c}{\textbf{CNN/DailyMail}} \\
\midrule
\textbf{Model} & \textbf{Type} & \textbf{R-1}$^{\uparrow}$ & \textbf{R-2}$^{\uparrow}$ & 
\textbf{R-L}$^{\uparrow}$ & \textbf{B.S.}$^{\uparrow}$ \\
\midrule
BART-L \citep{lewis2019bartdenoisingsequencetosequencepretraining} & N & 44.2 & 21.1 & 40.9 & 69.7 \\
RPE-BART-L & G & 44.5 & 21.2 & 41.3 & 69.4 \\
PRE-BART-L & G & 45.3 & 21.9 & 42.2 & 69.8 \\
\midrule
\multicolumn{6}{c}{\textbf{XSum}} \\
\midrule
\textbf{Model} & \textbf{Type} & \textbf{R-1}$^{\uparrow}$ & \textbf{R-2}$^{\uparrow}$ & 
\textbf{R-L}$^{\uparrow}$ & \textbf{B.S.}$^{\uparrow}$ \\
\midrule
BART-L \citep{lewis2019bartdenoisingsequencetosequencepretraining} & N & 45.14 & 22.27 & 37.25 & 73.3 \\
RPE-BART-L & G & 44.5 & 20.8 & 35.6 & 72.3 \\
PRE-BART-L & G & 45.2 & 21.3 & 36.4 & 72.7 \\
\bottomrule
\end{tabular}

\caption{ROUGE and BERTScore results on CNN/DailyMail and XSum.  
\textbf{Legend:} G = length from the reference summary is given;  
N = no explicit length control.}
\label{tab:scores_results}
\end{table}

\subsection*{Generalization to out-of-distribution Data }

\begin{figure}[!ht]
    \includegraphics[width=0.45\textwidth]{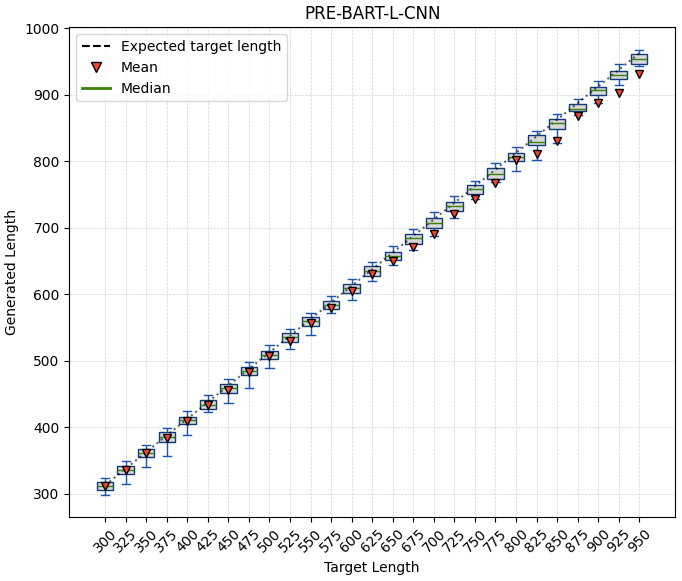}
    \caption{MAE by target-length bucket (25 tokens) for
PRE-BART-L on CNN/DailyMail when target lengths
above 300 tokens are requested.}
    \label{fig:pre_summaries_out_distrib}
\end{figure}

We have introduced the PRE method to achieve more robust generalization of length control across the full range of lengths that the model can handle. Figure~\ref{fig:pre_summaries_out_distrib} illustrates that PRE-BART-L is actually able to generate summaries with accurate length control, even when the requested output length exceeds 300 tokens.

To evaluate behavior along all manageable context windows, we also investigate a comparative analysis between PRE and RPE methods on 6,000 randomly selected examples from the CNN/Daily Mail test set. For each selected article, we randomly assigned a target summary length smaller than the article length and asked both models to generate a summary under this constraint. 
Table~\ref{tab:outlier_comparison} reports corresponding results, in terms of the percentage of produced outliers for which MAE greater than 20 tokens ($\%$ Out). We observe that the proportion of outliers is considerably smaller with PRE, highlighting its robustness in handling long and variable-length generation tasks.  PRE encodes a continuous ratio signal normalized between 0 and 1. This distinction is crucial for handling arbitrary user-specified output lengths. RPE is more sensitive to the distribution of target lengths in the training data. As shown in Appendix~\ref{sec:appendixC}, long target lengths are underrepresented in the original dataset. Consequently, models trained with RPE rarely encounter such cases during fine-tuning, thus their parameters are not adapted to accommodate very long summaries.

\begin{table}[!t]
\centering
\small
\setlength{\tabcolsep}{1pt}
\renewcommand{\arraystretch}{1}
\setlength{\tabcolsep}{4pt}
\begin{tabular}{cccc}
\toprule
\textbf{Bucket} & \textbf{\% Out RPE} $\downarrow$ & \textbf{\% Out PRE} $\downarrow$ & \textbf{Count} \\
\midrule
\bucket{300}{350}  & 10.8\% & 0.4\%  & 839 / 922 \\
\bucket{350}{400}  & 15.4\% & 1.0\%  & 705 / 809 \\
\bucket{400}{450}  & 22.3\% & 1.5\%  & 608 / 745 \\
\bucket{450}{500}  & 28.0\% & 3.4\%  & 453 / 629 \\
\bucket{500}{550}  & 37.0\% & 4.3\%  & 357 / 577 \\
\bucket{550}{600}  & 50.1\% & 5.4\%  & 245 / 490 \\
\bucket{600}{650}  & 51.4\% & 4.0\%  & 216 / 429 \\
\bucket{650}{700}  & 53.9\% & 7.4\%  & 179 / 374 \\
\bucket{700}{750}  & 64.8\% & 10.9\% & 124 / 351 \\
\bucket{750}{800}  & 70.3\% & 9.9\%  & 93  / 273 \\
\bucket{800}{850}  & 75.1\% & 9.0\%  & 76  / 242 \\
\bucket{850}{900}  & 81.3\% & 8.4\%  & 43  / 230 \\
\bucket{900}{950}  & 84.9\% & 8.9\%  & 35  / 214 \\
\bucket{950}{1000} & 95.8\% & 19.8\% & 4   / 65  \\
\bottomrule
\end{tabular}
\caption{Comparison of outlier rate between RPE-BART-L and PRE-BART-L across length buckets. The Count column reports, respectively, the number of non-outlier samples for each model.
}
\label{tab:outlier_comparison}
\end{table}

\subsection*{Question generation accomodation}
To demonstrate the strong generalization capability of our method across diverse neural text generation tasks, we conducted experiments on the SQuAD dataset. We fine-tune the BART model with both RPE and PRE for the question generation task, where the objective is to generate a relevant question given a context passage and its associated answer. As shown in Tables~\ref{tab:squad_results_len} and \ref{tab:squad_results_scores}, the PRE variant exhibits clear advantages: it enables highly accurate length-controlled question generation while preserving the semantic quality of the outputs. Moreover, our method achieves strong results across multiple evaluation metrics, including BLEU, ROUGE, and BERTScore.

\begin{table}[h!]
    \centering

    \renewcommand{\arraystretch}{1.1}
    
    \begin{tabular}{@{}lcc@{}}
        \toprule
        \textbf{Model} & \textbf{Type} & \textbf{MAE}$^{\scriptsize \downarrow}$ $\pm$ \textbf{SD} \\ 
        \midrule
        BART-L        & N & 3.12 $\pm$ 3.3 \\
        RPE-BART-L    & G & 0.8 $\pm$ 3.6 \\
        PRE-BART-L    & G & 0.0 $\pm$ 0.1 \\
        \bottomrule
    \end{tabular}

    \caption{Length control evaluation on SQuAD using MAE and standard deviation (SD). 
    \textbf{Legend:} G = reference length; N = no explicit length control. 
    Lower values indicate better control.}
    \label{tab:squad_results_len}
\end{table} 

\begin{figure}[!ht]
    \includegraphics[width=0.43\textwidth]{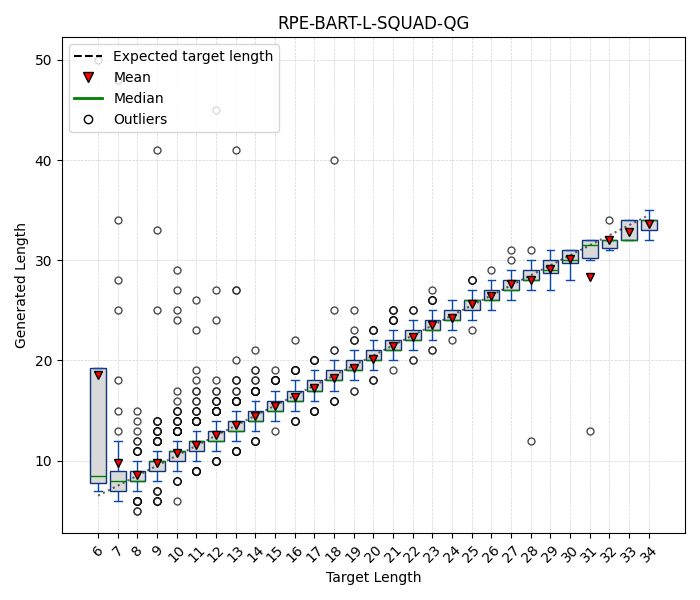}
    \caption{MAE by target-length bucket (1 tokens) for
RPE-BART-L on SQuAD dataset.}
    \label{fig:rpe_squad_len}
\end{figure}

\begin{figure}[!ht]
    \includegraphics[width=0.43\textwidth]{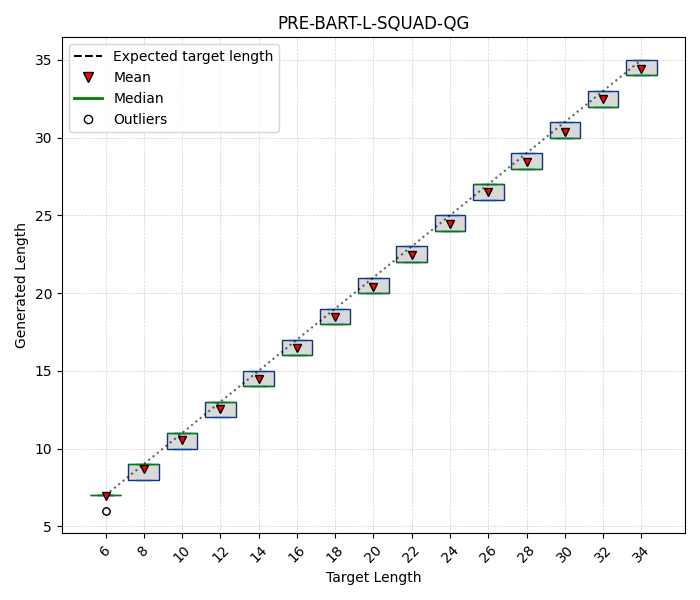}
    \caption{MAE by target-length bucket (2 tokens) for
PRE-BART-L on SQuAD dataset.}
    \label{fig:pre_squad_len}
\end{figure}

These experiments demonstrate the scalability of the PRE method to a broad range of text generation tasks. In particular, PRE provides more precise control over the target length than the RPE method (Figures~\ref{fig:rpe_squad_len} and \ref{fig:pre_squad_len}), especially near the extremities of the desired length distribution observed during training. This highlights the robustness and reliability of PRE for fine-grained length-controlled generation, even under demanding representation constraints.

\begin{table}[H]
\centering
\small
\setlength{\tabcolsep}{6pt}
\renewcommand{\arraystretch}{1.2}

\begin{tabular}{lccc}
\multicolumn{4}{c}{\textbf{SQuAD}} \\
\midrule
\textbf{Score} & \textbf{BART-L} & \textbf{RPE-BART-L} & \textbf{PRE-BART-L} \\
\midrule
BLEU$^{\uparrow}$  & 16.7 & 16.4 & 18.6\\
R-1$^{\uparrow}$  & 54.4 & 52.1 & 55.3\\
R-2$^{\uparrow}$  & 32 & 30.1 & 32.9 \\
R-L$^{\uparrow}$  & 50.1 & 47.7 & 50.8 \\
B.S.$^{\uparrow}$ & 75.2 & 72.6 & 74.8 \\
\bottomrule
\end{tabular}

\caption{BLEU, ROUGE and BERTScore results on the SQuAD dataset.}
\label{tab:squad_results_scores}
\end{table}

\section{Conclusion}
\label{sec:conclusion}

In this paper, we have introduced Progress Ratio Embeddings (PRE) as a novel and robust mechanism for length control in neural text generation. Unlike traditional Reverse Positional Embeddings (RPE), which rely on discrete countdown signals and struggle with generalization beyond the training distribution, PRE uses continuous impatience signals to encode progress in a more scalable way. Our experiments across different tasks generation highlight that PRE significantly improves length controllability while maintaining or enhancing text generation quality, according to classic metrics. Furthermore, PRE enables generalization to out-of-distribution target lengths, reducing the proportion of outliers and mean absolute error compared to RPE. Beyond providing length control in abstractive summarization \citep{ALMOHAIMEED2025100762}, PRE opens up new directions for fine-grained control over the generation process, enabling more progressive summarization while improving explainability.

\section{Limitations}
\label{sec:limitation}
Several limitations can be addressed in our work. Let us briefly discuss them. In this article, we have only describle our \textsc{PRE} method within an encoder-decoder architecture, which is not the prevailing trend of current decoder-only language models. A natural next step would be to evaluate the robustness of length control with the \textsc{PRE} method on decoder-only models and explore how this length control could facilitate step-by-step reasoning in content production, especially when given a progressively decreasing target length. Moreover, length control is not only relevant for summarization and question generation tasks but many other text-to-text tasks could benefit from this capability to regulate the generated length. Finally, an interesting direction for future research would be to explore length control within \textit{chain-of-thought} models. Being able to regulate the reasoning length could make such models more efficient at inference time, by preventing unnecessary over-generation and by allowing a deterministic adjustment of the reasoning depth.

\section{Ethical Considerations}
\label{sec:acknowledgment}
We have designed our methods to be simple to reproduce as they rely on publicly available pretrained language models. We acknowledge that any text generation system carries potential risks of producing biased or misleading content. To mitigate these risks, we restricted our experiments to widely used, publicly datasets whose content is considered as relatively safe for research purposes. Our work complies with the \textsc{ACL Ethics Policy}~\footnote{\url{https://www.aclweb.org/portal/content/acl-code-ethics}}.
\section{LLM Usage Declaration}

During the preparation of this manuscript, we used Large Language Models to assist with text clarity, grammar, and formulation, particularly in polishing the abstract and certain explanatory sentences. The scientific content, experimental design, results, and interpretations were entirely conceived and written by the authors. LLMs were not used to generate original ideas, proofs, or analyses; its contribution was limited to language refinement.

\section{Acknowledgements}
This work benefited from access to the HPC resources of GLICID \footnote{\url{https://glicid.fr/}} and resources of IDRIS \footnote{\url{https://www.genci.fr/}} under the allocation \texttt{2025–AD011016042} and \texttt{2026–AD011014032R3} made by GENCI.

\appendix
\section{Appendix : Stress signal plot for varying pulsations} 
\label{sec:appendixA}

We use trigonometric functions to represent the concept of a periodic impatience signal as:
\begin{equation*}
\label{eq:signal_impatience}
\begin{aligned}
s : \mathbb{R}^{+} \times [0,1] &\longrightarrow [-1,1]^2\\
(\omega,x) &\longmapsto \bigl(\cos(\omega x),\, \sin(\omega x)\bigr), 
\end{aligned}
\end{equation*}
where \(\omega\) denotes the pulsation of the periodic signal with a corresponding frequency \(F = \dfrac{\omega}{2\pi} \).

\begin{figure}[H] 
    \centering
    \includegraphics[width=0.47\textwidth]{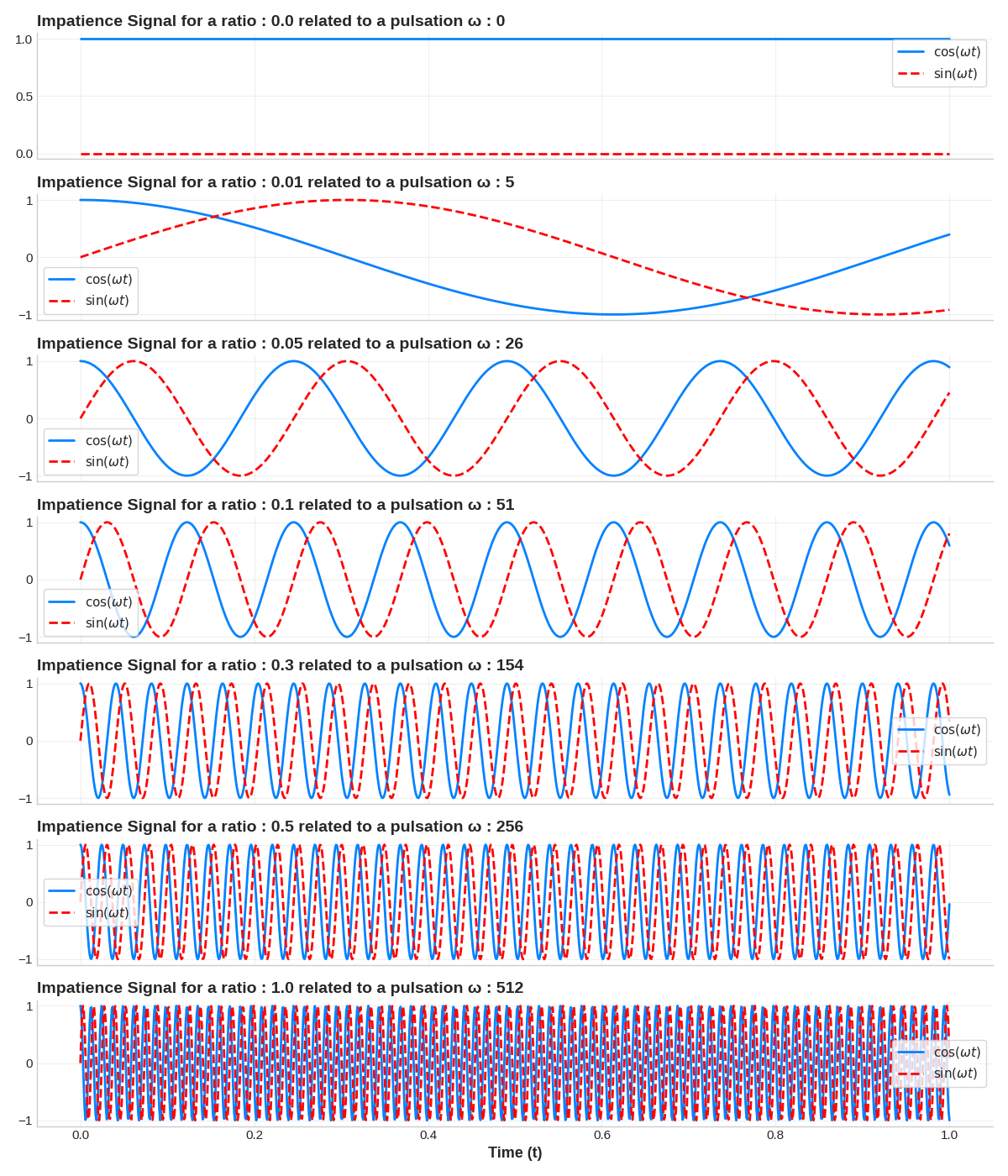}
    \caption{Impatience signal curves plot for differents pulsation $\omega$ based on $M=\dfrac{BART_{d_{model}}}{2} = 512$.}

    \label{fig:stress_signal}
\end{figure}

\section{Appendix : Illustration of the RPE method in relation to the decoder block input} 
\label{sec:appendixB}

\noindent
Reverse Positional Embeddings (RPE) extend the standard sinusoidal formulation \citep{NIPS2017_3f5ee243} by explicitly encoding the number of tokens remaining until the desired target length. 
Instead of encoding the absolute position $i$, RPE use the countdown index $(l-i)$, where $l$ is the target length. 

\begin{align}
RPE(i, 2k)   &= \sin\!\left(\frac{l-i}{10000^{2k/d_{model}}}\right), \\
RPE(i, 2k+1) &= \cos\!\left(\frac{l-i}{10000^{2k/d_{model}}}\right),
\end{align}

\noindent
where $d_{model}$ is the embedding dimension and $k$ indexes the dimensions. 
This design injects a discrete countdown signal into the decoder, enabling the model to condition generation explicitly on the remaining length.  

\begin{figure}[H]
    \centering
    \includegraphics[width=0.4\textwidth]{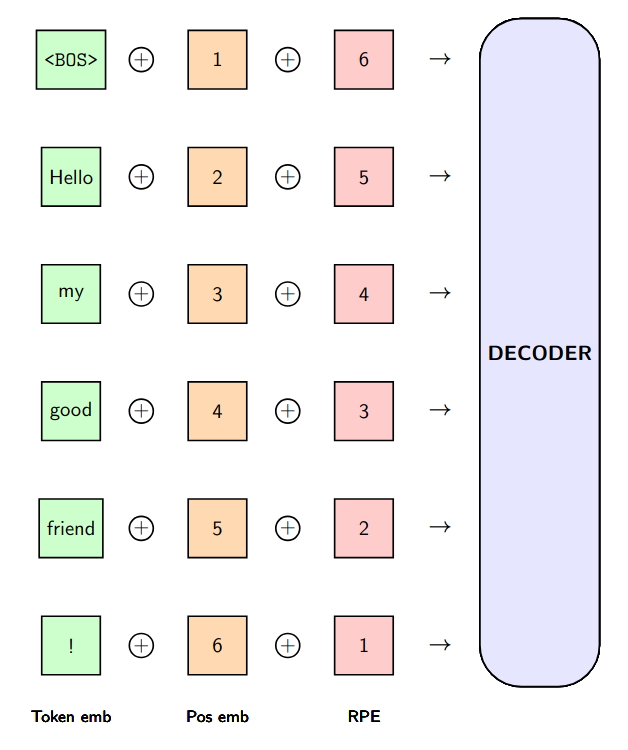} 
    \caption{Illustration of Reverse Positional Embeddings (RPE) added to token and positional embeddings before being passed to the decoder block.}
    \label{fig:RPE_schema}
\end{figure}

\section{Appendix : Summary lengths distribution throught datasets} 
\label{sec:appendixC}

\begin{figure}[H]
    \centering
    \includegraphics[width=0.5\textwidth]{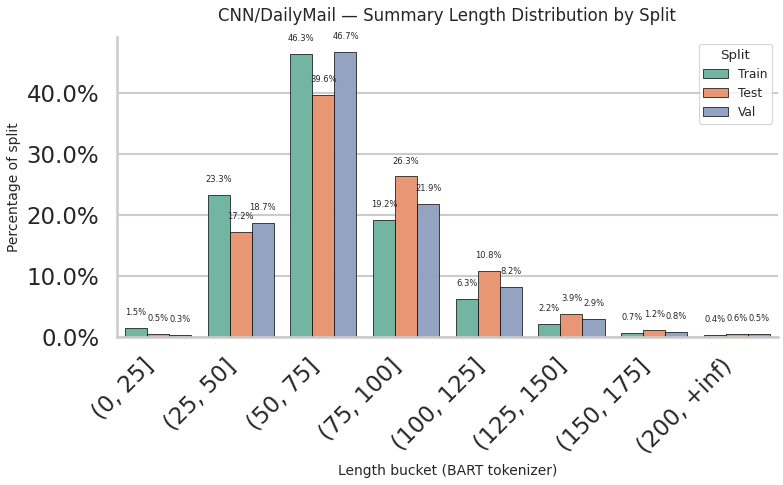} 
    \caption{Data visualisation over the CNN/DailyMail summary length distribution.}
    \label{fig:lenght_distrib_cnn_2}
\end{figure}

\begin{figure}[H]
    \centering
    \includegraphics[width=0.5\textwidth]{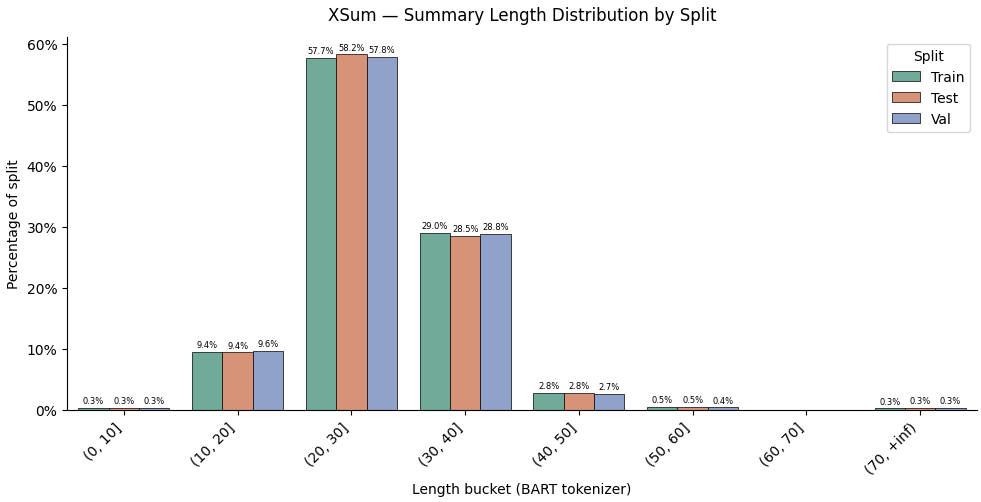} 
    \caption{Data visualisation over the XSum summary length distribution.}
    \label{fig:lenght_distrib_cnn_1}
\end{figure}

\section{Appendix : Results of the PRE method adaptation on the T5 encoder-decoder architecture} 
\label{sec:appendixD}

The \textsc{T5}-Large model \citep{raffel2023exploringlimitstransferlearning} treats every NLP task as text-to-text. Pre-trained on the C4 corpus with a span-corruption objective, it contains roughly 770M parameters and accepts inputs of up to 512 tokens. Developed by \citet{google}, this open-source model offers strong transferability across tasks and remains competitive for abstractive summarization. For this model $d_{model} = 512$.

\begin{table}[H]
    \centering
    \small
    \setlength{\tabcolsep}{6pt}
    \begin{tabular}{lccc}
        \toprule
        \textbf{Dataset} & \textbf{Model} & \textbf{Type} & \textbf{MAE}$^{\scriptsize \downarrow}$ $\pm$ \textbf{SD} \\
        \midrule
        \textbf{CNN/DM} 
            & T5-L        & N & 20.1 $\pm$ 17.9 \\
            & PRE-T5-L    & G & \textbf{7.9} $\pm$ 7.8 \\
        \midrule
        \textbf{XSum} 
            & T5-L        & N & 6.5 $\pm$ 5.4 \\
            & PRE-T5-L    & G & \textbf{1.9} $\pm$ 3.3 \\
        \bottomrule
    \end{tabular}
    \caption{Length control evaluation on CNN/DailyMail and XSum using MAE and standard deviation (SD). 
    \textbf{Legend:} G = guided by reference length; N = no explicit length control. Lower values indicate better control.}
    \label{tab:t5_length_control}
\end{table}

\begin{table}[H]
\centering
\small
\setlength{\tabcolsep}{1pt}
\renewcommand{\arraystretch}{1.1}

\begin{tabular}{lccccc}
\toprule
\multicolumn{6}{c}{\textbf{CNN/DailyMail}} \\
\midrule
\textbf{Model} & \textbf{Type} & \textbf{R-1}$^{\uparrow}$ & \textbf{R-2}$^{\uparrow}$ & 
\textbf{R-L}$^{\uparrow}$ & \textbf{B.S.}$^{\uparrow}$ \\
\midrule
T5-L \citep{raffel2023exploringlimitstransferlearning} & N & 42.50 & 20.68 & 39.75 & 68.2 \\
PRE-T5-L & G & 44.6 & 21.0 & 36.0 & 69.4 \\
\midrule
\multicolumn{6}{c}{\textbf{XSum}} \\
\midrule
\textbf{Model} & \textbf{Type} & \textbf{R-1}$^{\uparrow}$ & \textbf{R-2}$^{\uparrow}$ & 
\textbf{R-L}$^{\uparrow}$ & \textbf{B.S.}$^{\uparrow}$ \\
\midrule
T5-L \citep{stept2023_t5_large_xsum_cnn_summarization} & N & 36.77 & 14.69 & 30.06 & 68.9 \\
PRE-T5-L & G & 42.8 & 19.1 & 34.3 & 71.7 \\
\bottomrule
\end{tabular}

\caption{ROUGE and BERTScore results on CNN/DailyMail and XSum for T5 models.  
\textbf{Legend:} G = length from the reference summary is given;  
N = no explicit length control.}
\label{tab:t5_scores_results}
\end{table}

\section{Appendix: Reproduction of the LAAM Method}
\label{sec:appendixE}

The Length-Aware Attention Mechanism (LAAM) \citep{liuLAAMLengthControl2022} modifies the cross-attention of a seq2seq model by boosting the top-$\ell_t$ attended encoder tokens according to the remaining target length. This encourages the encoder–decoder attention to select information compatible with the desired output size. In our reproduction, LAAM is applied only to the last decoder layer of BART and fine-tuned on CNN/DailyMail and on SQuAD. Below we report the full set of scores obtained from our implementation. 

\begin{table}[H]
\centering
\small
\setlength{\tabcolsep}{4pt}
\renewcommand{\arraystretch}{1.1}

\begin{tabular}{lccccc}
\toprule
\textbf{Stat} & \textbf{R-1} & \textbf{R-2} & \textbf{R-L} & \textbf{B.S.} & \textbf{MAE} \\
\midrule
Mean & 44.2 & 21.1 & 41.1 & 69.6 & 17.5 \\
Std  & 12.5 & 13.6 & 12.6 & 6.9  & 16.2 \\
Min  & 0.0  & 0.0  & 0.0  & 42.4 & 0.0 \\
25\% & 36.0 & 11.6 & 32.7 & 65.4 & 5.0 \\
50\% & 44.0 & 18.8 & 40.5 & 69.6 & 10.0 \\
75\% & 51.9 & 27.9 & 48.5 & 74.0 & 21.0 \\
\bottomrule
\end{tabular}

\caption{LAAM performance on CNN/DailyMail.}
\label{tab:laam_cnn}
\end{table}

\begin{figure}[!ht]
    \includegraphics[width=0.4\textwidth]{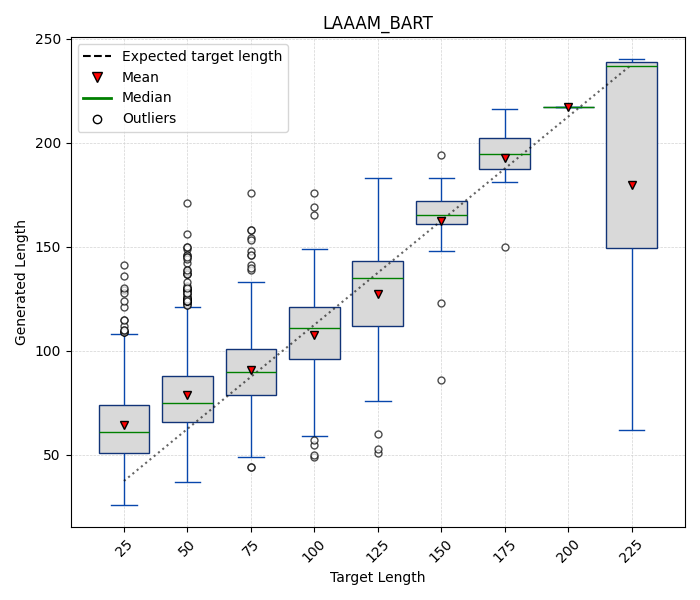}
    \caption{MAE by target-length bucket (25 tokens) for
LAAM-BART-L on CNN/DailyMail dataset.}
    \label{fig:laam_cnn}
\end{figure}

\begin{table}[H]
\centering
\small
\setlength{\tabcolsep}{4pt}
\renewcommand{\arraystretch}{1.1}

\begin{tabular}{lcccccc}
\toprule
\textbf{Stat} & \textbf{R-1} & \textbf{R-2} & \textbf{R-L} &
\textbf{BLEU} & \textbf{B.S.} & \textbf{MAE} \\
\midrule
Mean & 54.7 & 32.1 & 50.2 & 16.8 & 75.4 & 3.12 \\
Std  & 23.4 & 27.6 & 24.3 & 27.1 & 12.4 & 3.18 \\
Min  & 0.0  & 0.0  & 0.0  & 0.0  & 38.9 & 0.0 \\
25\% & 37.5 & 10.0 & 31.3 & 0.0  & 66.5 & 1.0 \\
50\% & 54.5 & 26.7 & 47.7 & 0.0  & 75.1 & 2.0 \\
75\% & 71.4 & 50.0 & 66.7 & 30.2 & 84.2 & 4.0 \\
\bottomrule
\end{tabular}

\caption{LAAM performance on SQuAD.}
\label{tab:laam_squad}
\end{table}

\begin{figure}[!ht]
    \includegraphics[width=0.4\textwidth]{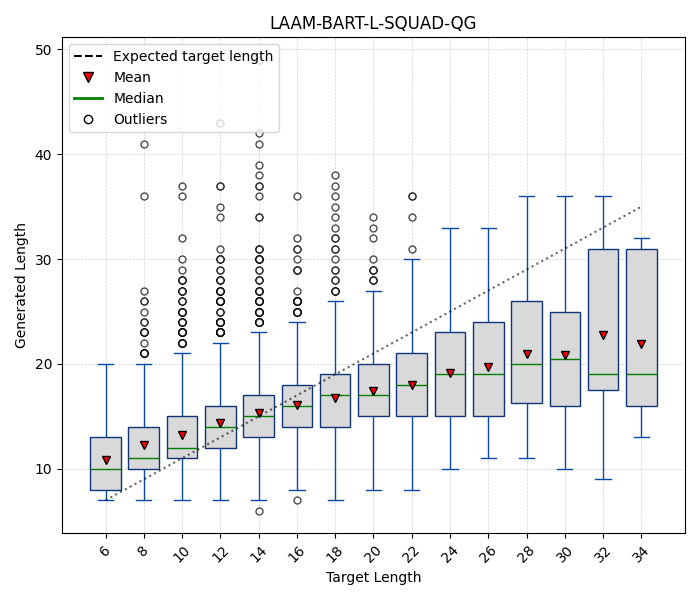}
    \caption{MAE by target-length bucket (2 tokens) for
LAAM-BART-L on SQuAD dataset.}
    \label{fig:laam_squad}
\end{figure}

\section{Appendix: Statistical Significance of Length Fidelity}
\label{sec:appendixF}

\noindent\textbf{Statistical Significance (Paired t-tests).}  
We performed paired Student’s \textit{t}-tests on example-level MAE values from the CNN/DailyMail test set to assess whether the improvements of PRE over existing baselines are statistically reliable.

\begin{itemize}
    \item PRE-BART vs.\ BART baseline: \\
    $t = -98.95,\; p < 10^{-300}$
    \item PRE-BART vs.\ RPE-BART: \\
    $t = -19.83,\; p = 4.8 \times 10^{-86}$
\end{itemize}

\noindent These results demonstrate that PRE yields highly statistically significant improvements in length fidelity.

\section{Appendix: Positioning with Length-Ratio Positional Encoding (LRPE)}
\label{sec:appendixG}

Prior approaches such as the length-ratio positional encoding (LRPE) introduced by \citet{takasePositionalEncodingControl2019}, which we initially cited as a source of inspiration proposes a formulation based on a ratio encoding during the decoding progress. We argue that, similarly to RPE, it still depends explicitly on the target length and we claim that this dependence hinders generalization, particularly for out-of-distribution target lengths at test time.

The LRPE formulation can be rewritten by introducing a progress ratio $r = \frac{pos}{len}$, similar to the one used in our approach:
\begin{align*}
&\text{LRPE}(pos, len, 2j) = \cos\!\left( \frac{pos}{len^{\frac{2j}{d_{\text{model}}}}} \right) \notag \\
&= \cos\!\left( r \cdot len^{1 - \frac{2j}{d_{\text{model}}}} \right), \\
&\text{LRPE}(pos, len, 2j+1) = \sin\!\left( \frac{pos}{len^{\frac{2j+1}{d_{\text{model}}}}} \right) \notag \\
&= \sin\!\left( r \cdot len^{1 - \frac{2j+1}{d_{\text{model}}}} \right).
\end{align*}
where $j$ indexes the embedding dimension (for a model of size $d_{\text{model}}$), $len$ denotes the target length, and $pos$ is the current decoding step.

Although LRPE incorporates a progress ratio, it remains explicitly dependent on $len$. During training, this implies that the inputs are not invariant to the target length. In particular, the same ratio yields different encodings depending on whether the expected sequence is long or short.

To empirically validate this hypothesis, we reproduced the LRPE method and scaled it to the BART-L architecture on the CNN/DailyMail dataset. As shown in Tables ~\ref{tab:lrpe_length_control} and ~\ref{tab:lrpe_scores}, LRPE yields less stable results and does not reach the same level of accuracy as PRE in our main experiments.

To evaluate robustness under strict length constraints outside the training distribution, we compute the Mean Absolute Error (MAE) between the generated length and randomly sampled target lengths that exceed the original reference length. Results are grouped into 25-token buckets.\footnote{The same random seed is used across all models for fair comparison.}

As shown in Table~\ref{tab:lrpe_outliers}, PRE-BART-L consistently achieves the lowest MAE across all length ranges, including extreme out-of-distribution regimes.

In contrast, RPE-BART-L exhibits a sharp and near-monotonic degradation as the target length increases, highlighting its limited extrapolation capability. LRPE-BART-L improves over RPE but still shows substantial error growth for longer sequences, indicating residual dependence on absolute length information.

\begin{table}[!t]
\centering
\small
\setlength{\tabcolsep}{1pt}
\renewcommand{\arraystretch}{1}
\setlength{\tabcolsep}{4pt}
\begin{tabular}{cccc}
\toprule
\textbf{Bucket} & \textbf{PRE-BART-L} & \textbf{RPE-BART-L} & \textbf{LRPE-BART-L} \\
\midrule
\bucket{200}{225} & 0.54  & 2.61   & 13.22 \\
\bucket{225}{250} & 0.63  & 3.18   & 16.91 \\
\bucket{250}{275} & 0.70  & 4.34   & 21.21 \\
\bucket{275}{300} & 0.87  & 8.60   & 26.72 \\
\bucket{300}{325} & 1.32  & 11.77  & 26.54 \\
\bucket{325}{350} & 1.19  & 10.71  & 30.95 \\
\bucket{350}{375} & 1.96  & 15.47  & 30.47 \\
\bucket{375}{400} & 1.58  & 20.69  & 34.04 \\
\bucket{400}{425} & 2.07  & 29.65  & 27.74 \\
\bucket{425}{450} & 3.20  & 37.02  & 26.51 \\
\bucket{450}{475} & 2.79  & 50.86  & 32.44 \\
\bucket{475}{500} & 2.96  & 57.29  & 43.14 \\
\bucket{500}{525} & 3.45  & 51.93  & 53.93 \\
\bucket{525}{550} & 4.03  & 58.24  & 58.28 \\
\bucket{550}{575} & 8.22  & 72.94  & 59.41 \\
\bucket{575}{600} & 7.91  & 94.44  & 46.09 \\
\bucket{600}{625} & 4.97  & 97.74  & 50.47 \\
\bucket{625}{650} & 6.83  & 113.09 & 45.10 \\
\bucket{650}{675} & 6.83  & 140.51 & 75.16 \\
\bucket{675}{700} & 12.78 & 143.27 & 86.23 \\
\bucket{700}{725} & 11.76 & 165.07 & 84.73 \\
\bucket{725}{750} & 8.48  & 170.40 & 117.84 \\
\bucket{750}{775} & 11.81 & 194.53 & 157.12 \\
\bucket{775}{800} & 7.33  & 164.31 & 141.47 \\
\bucket{800}{825} & 12.56 & 188.24 & 128.55 \\
\bucket{825}{850} & 7.60  & 242.80 & 126.77 \\
\bucket{850}{875} & 10.07 & 257.50 & 123.00 \\
\bucket{875}{900} & 16.88 & 266.51 & 137.03 \\
\bucket{900}{925} & 14.44 & 252.15 & 121.90 \\
\bucket{925}{950} & 9.49  & 329.41 & 105.84 \\
\bottomrule

\end{tabular}

\caption{MAE by target-length bucket in out-of-distribution settings. Lower values indicate better length control.}
\label{tab:lrpe_outliers}
\end{table}

\begin{table}[b!]
    \centering
    \small
    \renewcommand{\arraystretch}{1.1}
    \begin{tabular}{@{}lccc@{}}
        \toprule
        \textbf{Dataset} & \textbf{Model} & \textbf{Type} & \textbf{MAE} $\downarrow$ $\pm$ \textbf{SD} \\ 
        \midrule
        \multirow{4}{*}{\textbf{CNN/DM}} 
            & BART-L        & N & 19.2 $\pm$ 17 \\
            & LRPE-BART-L   & G & 2.19 $\pm$ 1.89 \\
            & RPE-BART-L    & G & 1.6 $\pm$ 3.6 \\
            & PRE-BART-L    & G & \textbf{0.5} $\pm$ 0.3 \\ 
        \bottomrule
    \end{tabular}
    \caption{Length control evaluation on CNN/DailyMail using mean absolute error (MAE) and standard deviation (SD). \textbf{Legend:} G = reference length provided; N = no explicit length control. Lower values indicate better control.}
    \label{tab:lrpe_length_control}
\end{table}

Overall, these results demonstrate that PRE provides significantly more stable and scalable length control, making it the most reliable approach for generation under unseen length constraints.

\begin{table}[h!]
\centering
\small
\setlength{\tabcolsep}{1pt}
\renewcommand{\arraystretch}{1.1}

\begin{tabular}{lccccc}
\toprule
\multicolumn{6}{c}{\textbf{CNN/DailyMail}} \\
\midrule
\textbf{Model} & \textbf{Type} & \textbf{R-1}$^{\uparrow}$ & \textbf{R-2}$^{\uparrow}$ & 
\textbf{R-L}$^{\uparrow}$ & \textbf{B.S.}$^{\uparrow}$ \\
\midrule
BART-L & N & 44.2 & 21.1 & 40.9 & 69.7 \\
LRPE-BART-L & G & 43.2 & 20.1 & 40.2 & 68.7 \\
RPE-BART-L & G & 44.5 & 21.2 & 41.3 & 69.4 \\
PRE-BART-L & G & \textbf{45.3} & \textbf{21.9} & \textbf{42.2} & \textbf{69.8} \\
\midrule
\end{tabular}
\caption{ROUGE and BERTScore results on CNN/DailyMail.  
}
\label{tab:lrpe_scores}
\end{table}

\section{Appendix: Evaluation with LLM-as-a-Judge (FineSurE)}
\label{sec:appendixH}

We conducted additional experiments using an LLM-as-a-judge evaluation metric, namely \textit{FineSurE} (Fine-grained Summarization Evaluation) \cite{song2024finesurefinegrainedsummarizationevaluation}. As the reference evaluator, we used \texttt{GEMMA-3-12B-Instruct} \citep{gemma_2025}\footnote{\url{https://huggingface.co/google/gemma-3-12b-it}}. Using the CNN/DailyMail test set, for each reference document--summary pair $(A, S)$, we computed the difference between the reference faithfulness score $\text{FineSurE}_{\text{Faithfulness}}(A, S)$ and the generated faithfulness score $\text{FineSurE}_{\text{Faithfulness}}(A, S^*)$, where $S^*$ denotes the summary generated by each model (BART-L, RPE-BART-L, PRE-BART-L).

We then measured the absolute difference:
\[
\Delta_{\text{Faith}} = \left| \text{Faithfulness}(A, S) - \text{Faithfulness}(A, S^*) \right|,
\]
where lower values indicate that the generated summary is closer to the reference in terms of faithfulness. Our results show that PRE-BART-L produces faithfulness scores that are substantially closer to the reference compared to the two baseline models.

\begin{table}[h!]
    \centering
    \small
    \renewcommand{\arraystretch}{1.1}
    \begin{tabular}{@{}lcc@{}}
        \toprule
        \textbf{Dataset} & \textbf{Model} & $\boldsymbol{\Delta_{\text{Faith}}}$ $\downarrow$ $\pm$ \textbf{SD} \\ 
        \midrule
        \multirow{3}{*}{\textbf{CNN/DM}} 
            & BART-L        & 0.12 $\pm$ 0.25 \\
            & RPE-BART-L    & 0.09 $\pm$ 0.26 \\
            & PRE-BART-L    & \textbf{0.01} $\pm$ 0.28 \\ 
        \bottomrule
    \end{tabular}
    \caption{Faithfulness distance to the reference using FineSurE (lower is better).}
    \label{tab:finesure_faithfulness}
\end{table}

\section{Appendix : Density comparison of summary length distributions} 
\label{sec:appendixI}

The PRE method produces summaries whose length distribution closely matches the true distribution observed in the reference data. 

\begin{figure}[!h]
\centering

\begin{subfigure}[t]{0.45\textwidth}
    \centering
    \includegraphics[width=\linewidth]{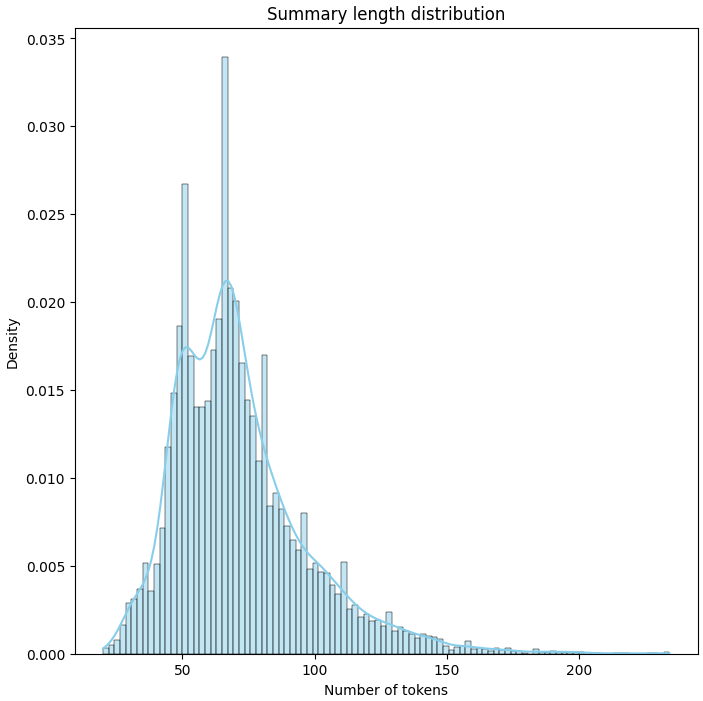}
    \caption{Reference length distribution.}
    \label{fig:distrib_cnn_target_length}
\end{subfigure}
\hfill
\begin{subfigure}[t]{0.45\textwidth}
    \centering
    \includegraphics[width=\linewidth]{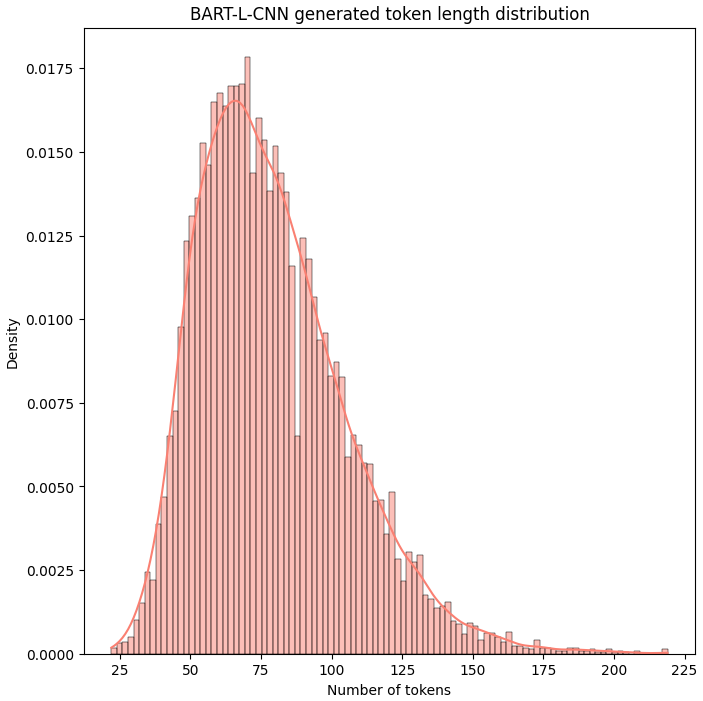}
    \caption{BART-L length distribution.}
    \label{fig:distrib_bart_cnn_target_length}
\end{subfigure}
\hfill
\begin{subfigure}[t]{0.45\textwidth}
    \centering
    \includegraphics[width=\linewidth]{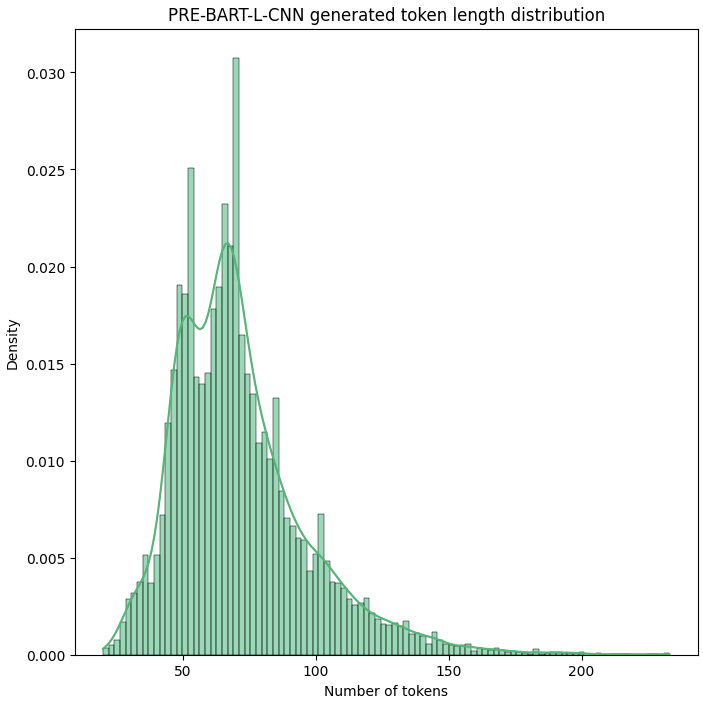}
    \caption{PRE-BART-L length distribution.}
    \label{fig:distrib_pre_bart_cnn_target_length}
\end{subfigure}

\caption{Density comparison of summary length distributions on CNN/DailyMail.}
\label{fig:three-wide-cnn-length-distrib}
\end{figure}

\end{document}